\documentclass[10pt,twocolumn,letterpaper]{article}

\usepackage{cvpr}

\usepackage[table]{xcolor}
\usepackage{adjustbox}
\usepackage{makecell}
\usepackage{booktabs}
\usepackage{multicol}
\usepackage{multirow}
\usepackage{pifont}
\usepackage[most]{tcolorbox}

\providecommand{\cmark}{}
\providecommand{\xmark}{}
\renewcommand{\cmark}{{\color[HTML]{2E8B57}\ding{51}}}
\renewcommand{\xmark}{{\color[HTML]{D0021B}\ding{55}}}

\tcbset{
    colback=gray!3,
    colframe=gray!60,
    arc=4pt,
    boxrule=0.5pt,
}

\definecolor{cvprblue}{rgb}{0.21,0.49,0.74}
\usepackage[pagebackref,breaklinks,colorlinks,allcolors=cvprblue]{hyperref}

\def\paperID{11044}
\def\confName{CVPR}
\def\confYear{2026}

\title{Extending Embodied Question Answering from Perception to Decision}

\author{
Xicheng Gong$^{1,2}$ \quad Qiwei Li$^{1}$ \quad Peiran Xu$^{1}$ \quad Yadong Mu$^{1,*}$\\
$^{1}$Peking University \quad $^{2}$XYZ Embodied AI\\
{\tt\small gongxicheng@stu.pku.edu.cn \quad lqw@pku.edu.cn \quad  xpr820@pku.edu.cn  \quad myd@pku.edu.cn}\\
{\tt\small $^{*}$Corresponding author}
}

\begin{document}
\maketitle
\vspace{-8mm}
\begin{abstract}
Embodied Question Answering (EQA) connects perception, reasoning, and interaction within embodied environments. However, existing datasets and benchmarks remain fragmented, each focusing on a limited subset of reasoning skills such as spatial understanding or procedural reasoning, without offering a unified large-scale framework for comprehensive evaluation. We present \textbf{EQA-Decision}, a large-scale embodied QA dataset that systematically covers four complementary dimensions of embodied reasoning: static scene construction, spatial understanding, task dynamics reasoning, and instant decision. The dataset contains over four million question–answer pairs with hierarchical annotations across diverse embodied scenarios. In addition, we develop \textbf{RoboDecision}, a strong baseline model aligned with the \textbf{EQA-Decision Benchmark}, providing a unified framework that jointly evaluates perception, reasoning, and action-level decision-making in embodied environments. Results demonstrate that \textbf{EQA-Decision} effectively benchmarks and enhances VLM capabilities in spatial and interaction reasoning, providing a solid foundation for advancing  embodied intelligence research.
\end{abstract}
\begin{figure*}[t]
    \centering
    \includegraphics[width=1\linewidth]{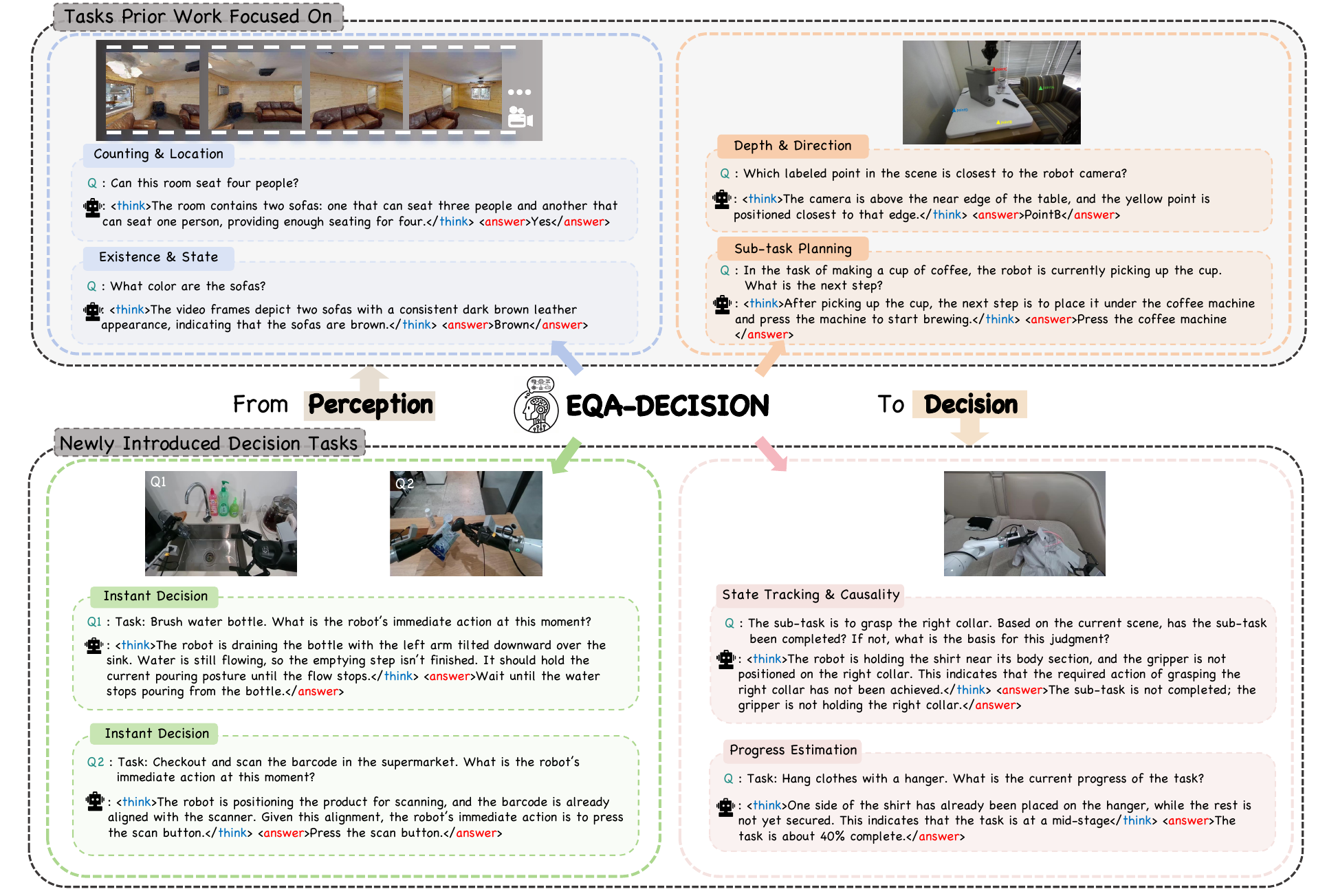}
    \caption{
        \textbf{Overview of EQA-Decision.}
        Prior embodied QA datasets and benchmarks mainly target perception-oriented tasks, where models focus on describing what is visible.
EQA-Decision extends this scope with decision-centric tasks that require understanding spatial and temporal context, tracking state changes, and performing reasoning within dynamic task processes.
    }
    \label{fig:intro_overview}
\end{figure*}    

\section{Introduction}
\label{sec:intro}

Recent advances in multimodal large language models (MLLMs) have greatly accelerated
the development of embodied intelligence. By integrating visual perception with
language reasoning, MLLMs have shown strong generalization across grounded tasks
such as visual question answering, spatial reasoning, and instruction following.
Models like GPT-4V~\cite{achiam2023gpt}, Gemini-2.5~\cite{comanici2025gemini}, and Qwen2.5-VL~\cite{bai2025qwen25vltechnicalreport} exhibit powerful multimodal understanding
and reasoning abilities, marking an important step toward unifying perception,
cognition, and action within a single framework.

Building on these advances, researchers have increasingly explored how to adapt MLLMs to robotic and embodied environments, 
which has led to a growing number of datasets and benchmarks for evaluating perception and reasoning in realistic settings. 
Recent efforts ~\cite{sermanet2024robovqa,chen2025robo2vlm,azzolini2025cosmos} concentrate on spatial understanding and task planning, 
while large-scale egocentric collections~\cite{grauman2022ego4d,damen2018scaling} provide diverse multimodal observations of human–object interactions. 
Complementary to these datasets, benchmarks including~\cite{das2018embodied,islam2023eqa,majumdar2024openeqa} aim to assess embodied reasoning and grounding capabilities through standardized evaluation protocols. 
However, existing datasets and benchmarks for embodied question answering remain limited in scope. 
Most focus on static perception or isolated reasoning skills such as spatial grounding and planning reasoning, 
while neglecting the decision-making process that unfolds as an agent interacts with a dynamic environment. 
Without explicit modeling of how perception, reasoning, and action evolve over time, 
current benchmarks cannot capture the instant decisions crucial for real embodied intelligence.

To address these limitations, we present \textbf{EQA-Decision}, a unified large-scale embodied question answering dataset that extends embodied reasoning from static perception to dynamic interaction. The dataset contains over 4 million multimodal question--answer pairs collected from both simulated and real-world sources, making it one of the largest embodied QA datasets to date. As shown in Fig.~\ref{fig:intro_overview}, in contrast to prior datasets that emphasize a single capability such as spatial reasoning or procedural planning, EQA-Decision encompasses a wide range of embodied scenarios and is organized into four major reasoning modules, namely static scene construction, spatial understanding, task dynamics reasoning, and instant decision-making, which are further divided into nine sub-tasks. Moreover, we introduce new task formulations for progress analysis and context-aware instant decision, enabling evaluation of agents’ ability to reason over temporal dynamics and adapt their actions in real time. This comprehensive coverage transforms embodied QA from static perception and spatial understanding toward a more temporally grounded and decision-oriented paradigm, bridging the gap between perception, reasoning, and action in embodied intelligence.

Building on this dataset, we further establish the \textbf{EQA-Decision Benchmark}, 
a unified evaluation suite designed to assess embodied reasoning across six complementary dimensions. 
The benchmark defines standardized task protocols and scoring metrics, enabling consistent and reproducible evaluation of perception, reasoning, and decision-making capabilities. This unified framework enables consistent evaluation of both open-source and closed-source multimodal models, including Qwen3-VL, Gemini-2.5-pro~\cite{comanici2025gemini}, and GPT-5, providing a clear basis for future embodied intelligence research.

We also introduce \textbf{RoboDecision}, a baseline model built upon Qwen3-VL-8B-Instruct. Through multi-stage SFT, CoT-SFT, and GRPO~\cite{shao2024deepseekmath} training with a hybrid reward that evaluates reasoning, correctness, and visual grounding, RoboDecision learns to better connect perception with decision-making. This decision-aware optimization enhances spatial grounding, temporal understanding, and action-level reasoning, making RoboDecision a strong unified baseline for embodied reasoning research.



\begin{table*}[t]
\centering
\small
\setlength{\tabcolsep}{2pt}   
\renewcommand{\arraystretch}{1.0}
\begin{adjustbox}{max width=\textwidth}
\begin{tabular}{l|
cc|ccc|ccc|c|c}
\toprule
\textbf{Dataset} &
\multicolumn{2}{c|}{\textbf{Static Scene Construction}} &
\multicolumn{3}{c|}{\textbf{Spatial Understanding}} &
\multicolumn{3}{c|}{\textbf{Task Dynamics Reasoning}} &
\textbf{Instant Decision} &
\textbf{CoT Annotation} \\

\cmidrule(lr){2-3}\cmidrule(lr){4-6}\cmidrule(lr){7-9}\cmidrule(lr){10-10}\cmidrule(lr){11-11}
& \makecell{Existence\\\&State}
& \makecell{Counting\\\&Location}
& \makecell{Depth\\\&Direction}
& \makecell{Grounding\\\&Referring}
& \makecell{Affor-\\dance}
& \makecell{Sub-task\\Planning}
& \makecell{State Tracking \\ \&Causality}
& \makecell{Progress\\Estimation}
& \makecell{Instant Decision}
& \makecell{CoT Annotation}  \\

\midrule

\textbf{Robo2VLM~\cite{chen2025robo2vlm}}     & \xmark & \xmark & \cmark & \xmark & \xmark & \cmark & \cmark & \xmark & \xmark & \xmark \\
\textbf{RoboVQA~\cite{sermanet2024robovqa}}      & \xmark & \xmark & \xmark & \xmark & \xmark & \cmark & \cmark & \xmark & \xmark & \xmark \\
\textbf{ShareRobot~\cite{ji2025robobrain}}     & \xmark & \xmark & \xmark & \cmark & \cmark & \cmark & \cmark & \xmark & \xmark & \xmark  \\

\rowcolor{cyan!10}
\textbf{EQA-Decision} &
\textbf{\cmark} & \textbf{\cmark} &
\textbf{\cmark} & \textbf{\cmark} & \textbf{\cmark} &
\textbf{\cmark} & \textbf{\cmark} & \textbf{\cmark} &
\textbf{\cmark} &\textbf{\cmark}\\
\bottomrule
\end{tabular}
\end{adjustbox}

\caption{\textbf{Comparison of embodied QA across reasoning dimensions.}
Columns indicate whether each dataset supports the corresponding task type; the last column marks the availability of chain-of-thought (CoT) annotations.
}
\label{tab:dataset_comparison}
\end{table*}

\section{Related Work}
\label{sec:related work}

\paragraph{Large-Scale EQA Datasets and Benchmarks} The Embodied Question Answering (EQA) task, first proposed by~\cite{das2018embodied}, integrates visual perception, navigation, and language reasoning in 3D environments~\cite{wu2018building,kolve2017ai2,dai2017scannet,ramakrishnan2021habitat}. Early datasets~\cite{das2018embodied,wijmans2019embodied,yu2019multi} emphasized goal-driven navigation with template-based questions in simulated worlds~\cite{kolve2017ai2,wu2018building}. Later works shifted toward spatial scene understanding. Subsequent research~\cite{azuma2022scanqa,ma2022sqa3d,dorbala2024house,wu2024noisyeqa,ren2024explore,majumdar2024openeqa} gradually moved beyond navigation toward 3D spatial and situational reasoning, expanding the embodied QA paradigm to real-world scenes that demand geometric understanding, object–relation reasoning, and interpretation of dynamic state changes within complex environments~\cite{ramakrishnan2021habitat,dai2017scannet}.

Recent efforts have further extended EQA from navigation and scene reasoning to embodied manipulation and task-level understanding. RoboVQA~\cite{sermanet2024robovqa} introduces video-based question answering to capture grasping and collaborative interactions, while Robo2VLM~\cite{chen2025robo2vlm} leverages real-world teleoperated robot trajectories to build large-scale embodied QA datasets. By incorporating proprioceptive and control-related signals such as gripper states and action phases, these works bridge perception and action, expanding EQA from scene-centric reasoning to action-centric embodied understanding.

\paragraph{MLLMs for Embodied Manipulation and Planning} Recent advances in multimodal large language models (MLLMs) have shifted embodied intelligence from passive perception toward active task understanding and manipulation planning. Early vision-language-action (VLA) systems~\cite{driess2023palm,brohan2022rt,zitkovich2023rt,kim2024openvla,shi2025hi,team2024octo}  demonstrated that large-scale alignment of vision, language, and action can yield language-conditioned control. Building upon these foundations, more recent works further incorporate action reasoning and lightweight world modeling to support structured decision-making in embodied contexts \cite{zhao2025cot,li2025simplevla,cen2025worldvla,zhong2025flowvla,zhang2025dreamvla}. However, most approaches still emphasize short-horizon decisions or direct input-to-action mappings, leaving long-term planning, causal reasoning, and adaptive replanning underexplored in complex real-world environments.

Complementary to these efforts, another line of research builds robot-centric embodied models that couple MLLM reasoning with physical control and domain-specific knowledge. Early agents \cite{chen2023robogpt,mu2023embodiedgpt}  adopt chain-of-thought based planning to decompose instructions into subgoals and connect them with low-level controllers. More recent systems \cite{ji2025robobrain,team2025robobrain,luo2025visual,azzolini2025cosmos} integrate structured world models and language-to-motion adapters, enabling robots to bridge abstract reasoning with real-world execution. Together, these specialized models highlight a transition from general VLA architectures to domain-adaptive embodied intelligence, where robots reason about affordance, feasibility, and task progression beyond static perception.

\section{Dataset Construction}
\label{sec:dataset}

\subsection{Overview}
\label{sec:overview}

\begin{figure*}[t]
    \centering
    \includegraphics[width=\linewidth]{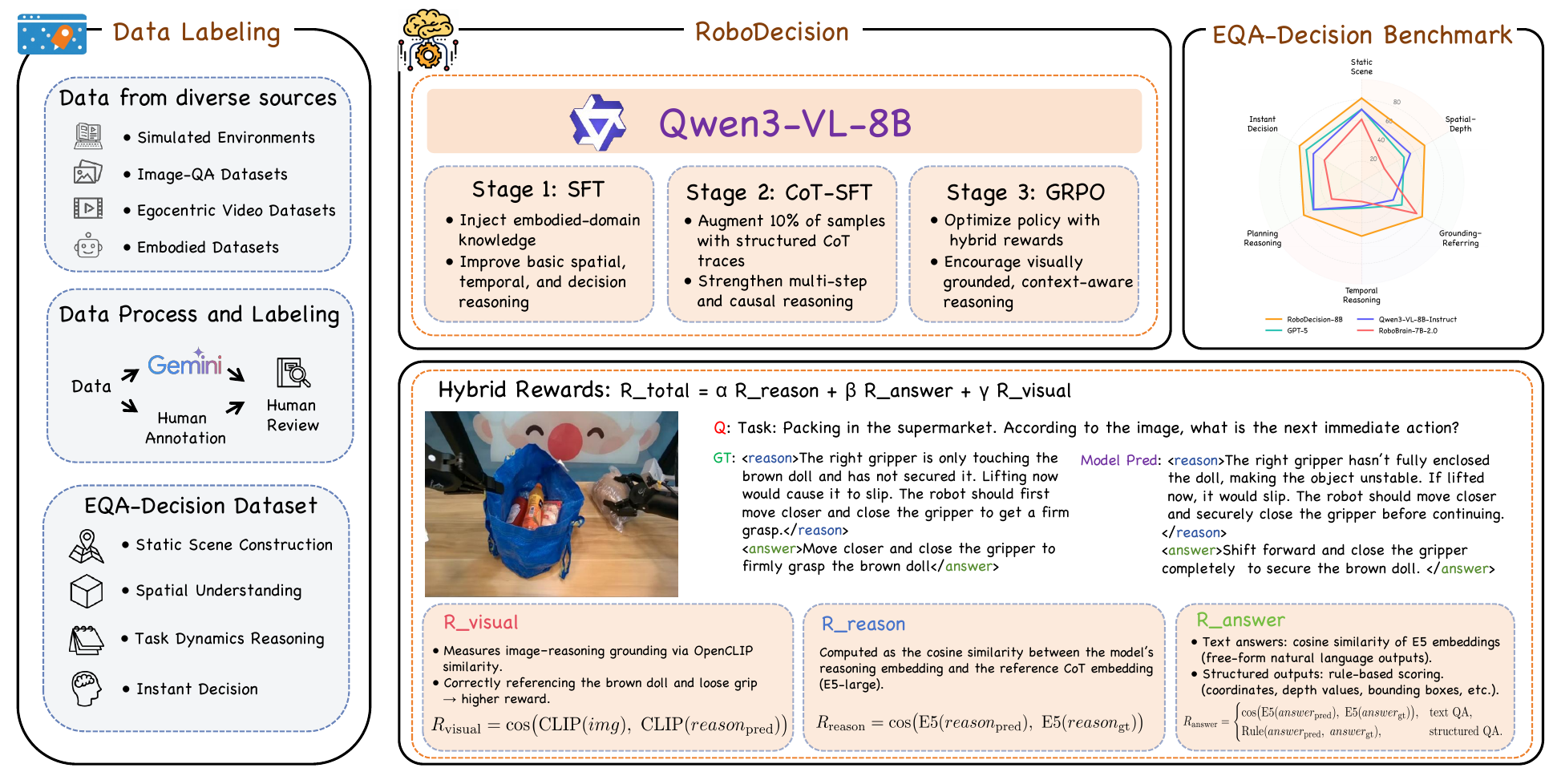}
    \caption{
        \textbf{Overall framework of EQA-Decision and RoboDecision.}
        We combine Gemini assisted annotation with human verification to process data from multiple sources, producing the EQA-Decision dataset with structured embodied reasoning tasks.
Building on this dataset, the model is trained in three successive stages: SFT, CoT-SFT and GRPO, with a hybrid reward applied throughout training to improve reasoning, answer accuracy and visual grounding.
The trained model is then evaluated on six types of embodied reasoning tasks in the EQA-Decision benchmark, demonstrating the breadth of the dataset and the effectiveness of the RoboDecision approach.
    }
    \label{fig:main}
\end{figure*}

Existing embodied QA datasets mainly focus on static or non-temporal tasks, which limits their ability to assess higher-level reasoning involving temporal understanding and adaptive decision-making. 
To overcome these limitations, we construct the EQA-Decision dataset, a large-scale and hierarchically organized resource that extends embodied QA from perception to decision. 
It contains more than four million multimodal question--answer pairs collected from diverse visual and embodied sources covering both simulated and real environments. 
The dataset is organized into four major reasoning modules: Static Scene Construction, Spatial Understanding, Task Dynamics Reasoning, and Instant Decision, together encompassing nine fine-grained sub-tasks. 
As shown in Table~\ref{tab:dataset_comparison}, EQA-Decision provides a more comprehensive and balanced coverage of embodied reasoning dimensions, including progress estimation and instant decision, which are not addressed in previous datasets.

\subsection{Data Sources}
\label{sec:sources}
\textbf{Simulated Environments}  
We build a diverse and controllable set of embodied trajectories using three indoor simulation platforms: HM3D~\cite{ramakrishnan2021habitat}, ScanNet~\cite{dai2017scannet}, and AI2-THOR~\cite{kolve2017ai2}. 
ScanNet~\cite{dai2017scannet} was collected by human operators scanning indoor scenes with RGB-D cameras. 
We use the first thirty seconds of each trajectory to construct realistic human-guided paths that serve as part of our dataset. 
Based on HM3D~\cite{ramakrishnan2021habitat}, we generate navigation trajectories by sampling start and goal locations and computing curved paths over ten meters to ensure diverse spatial coverage. 
In AI2-THOR~\cite{kolve2017ai2}, we replay task environments within the ALFRED~\cite{shridhar2020alfred} framework using the CAPEAM~\cite{kim2023context} agent, recording complete robotic trajectories in household scenes along with synchronized visual and semantic information.
In total, we collect around twenty thousand trajectories from these simulated environments, covering kitchen, office, and living-room scenes.

\textbf{Image-QA Datasets}  
We incorporate three image-QA sources to enhance visual grounding and embodied reasoning: Bunny-695K~\cite{he2024efficient}, RoboVQA~\cite{sermanet2024robovqa}, and RoboPoint~\cite{yuan2024robopoint}. Based on Bunny-695K~\cite{he2024efficient} , we refine question types and add spatial and depth annotations to improve object level understanding. 
We use RoboVQA~\cite{sermanet2024robovqa} as a task-related data source and further reannotate and process it to align with our embodied reasoning framework. 
Additionally, we sample examples from RoboPoint~\cite{yuan2024robopoint}, which provides affordance-oriented QAs with keypoint level spatial cues. 

\textbf{Egocentric Video Datasets}  
We include two large-scale egocentric video datasets, Ego4D~\cite{grauman2022ego4d} and EPIC-KITCHENS-100~\cite{Damen2021PAMI}, to capture realistic human–object interactions and temporal dynamics. 
Ego4D provides diverse first-person recordings of daily activities, offering natural visual context for temporal reasoning. 
EPIC-KITCHENS-100 focuses on kitchen tasks with detailed action and object annotations, enriching procedural and action understanding. 
We extract representative frames and temporal clips to support embodied reasoning in egocentric settings.

\textbf{Embodied Datasets}  
To bridge simulation and real-world manipulation, we incorporate large-scale embodied datasets capturing multimodal robot trajectories in diverse physical environments.  
From the Open X-Embodiment corpus, we curate representative subsets ~\cite{zhu2023fanuc, shah2023mutex, mandlekar2019scaling, lynch2023interactive, walke2023bridgedata, jang2021bc, liu2022robot, nasiriany2022sailor, cui2022play, dass2023jacoplay,bahl2023affordances,vogel2020edan}, covering multiple robot types and manipulation skills including pick-and-place, insertion, and tool use. Each trajectory contains synchronized RGB observations, proprioceptive states, and action annotations for accurate perception–action grounding.  
We also integrate long-horizon teleoperated data from AgiBot World ~\cite{bu2025agibot}, which includes over 80 daily manipulation skills across realistic household, industrial, and office scenes.  
Detailed descriptions of all data sources are available in Supplementary Material.
\subsection{Annotation and Processing Pipeline}
\label{sec:pipeline}

\textbf{Static Scene Construction}  
This module focuses on basic scene understanding tasks including object existence, state, counting, and location. 
For trajectories from HM3D~\cite{ramakrishnan2021habitat} and ScanNet~\cite{dai2017scannet}, we first use Gemini-2.5-pro~\cite{comanici2025gemini} to summarize visible objects and their attributes, covering existence, state, color, and other key properties. 
These summaries are combined with corresponding RGB frames to generate annotations for the two sub-tasks, existence and state as well as counting and location. 
For ALFRED~\cite{shridhar2020alfred}, we transform execution metadata that record object states and spatial relations into structured QA annotations. 
Filtered samples from Bunny-695K~\cite{he2024efficient} are also included to enrich this subset and form a complete corpus for static scene reasoning.

\textbf{Spatial Understanding}  
 Spatial reasoning in this module is explored from three complementary perspectives: depth and direction, grounding and referring, and affordance.
We sample representative RGB frames from Bunny-695K~\cite{he2024efficient} and Open X-Embodiment~\cite{o2024open}, and estimate depth maps using ZoeDepth~\cite{https://doi.org/10.48550/arxiv.2302.12288}. 
From each frame, we extract geometric cues including relative distance, orientation, and object depth. 
We employ Gemini-2.5-pro~\cite{comanici2025gemini} to generate factual question–answer pairs conditioned on the RGB–depth inputs, covering topics such as relative position, depth comparison, and spatial arrangement. 
For grounding and referring tasks, we adopt object level and region level samples from RoboPoint~\cite{yuan2024robopoint}, which link textual descriptions with corresponding image regions. 
We further annotate affordance areas in Open X-Embodiment~\cite{o2024open} frames, labeling each region with a concise description that indicates its manipulability, for example graspable, pushable, or openable. 

\textbf{Task Dynamics Reasoning}  
The module is organized into three components that capture temporal and causal relationships in embodied tasks: sub-task planning, state tracking, and progress estimation.  
Temporally annotated datasets, including Ego4D~\cite{grauman2022ego4d}, EPIC-KITCHENS-100~\cite{Damen2021PAMI}, AgiBot~\cite{bu2025agibot}, and Open X-Embodiment~\cite{o2024open}, are used to construct samples for sub-task planning. Frames around annotated temporal boundaries are extracted to construct question-answer pairs related to contextual planning, remaining steps, and future predictions. Question formulations are expanded using Gemini-2.5-pro~\cite{comanici2025gemini} to increase the linguistic diversity while maintaining temporal consistency. State tracking and causal reasoning are processed with the same framework, aiming to determine which sub-task the robot is currently executing, whether it has been successfully completed, and why.
Around each annotated sub-task boundary, we uniformly sample frames to construct balanced positive and negative examples, and use Gemini-2.5-pro~\cite{comanici2025gemini} to infer the underlying reasoning behind these judgments.  
For progress estimation, trajectories from AgiBot~\cite{bu2025agibot} and Open X-Embodiment~\cite{o2024open} are segmented into motion phases by velocity and directional variations. Normalized progress ratios are computed from frame indices.  

\textbf{Instant Decision}
This module focuses on modeling the robot’s real-time decision-making process in dynamic embodied environments. 
We use densely annotated video datasets including Ego4D~\cite{grauman2022ego4d}, EPIC-KITCHENS-100~\cite{Damen2021PAMI}, AgiBot~\cite{bu2025agibot}, and Open X-Embodiment~\cite{o2024open}. 
For each dataset, continuous action segments are selected where the interval between adjacent annotations does not exceed five seconds, ensuring fine-grained temporal continuity.  

Between every two consecutive annotations, we randomly sample intermediate frames that represent transient decision states. 
For each sampled frame, Gemini-2.5-pro~\cite{comanici2025gemini} is provided with neighboring annotations and the overall task description, allowing it to summarize key contextual details including the current step, completion status, and spatial relations between the agent and the relevant objects. 
These summaries, together with the corresponding visual inputs, are then used to construct question-answer pairs that capture context-aware decisions and next-action predictions.  
To ensure annotation quality, we conduct random human verification on sampled subsets to remove inconsistent or ambiguous examples. 
This process produces a dense corpus of instant-decision questions that reflect fine-grained perception–action reasoning and real-time adaptability in embodied interaction.

For all annotation procedures, we conduct both manual reviews and user evaluations to assess labeling quality. 
Detailed descriptions of the annotation protocol and evaluation setup are provided in supplementary material.

\subsection{Data Statistics and Distribution}

\begin{figure}[t]
    \centering
    \includegraphics[width=0.9\linewidth]{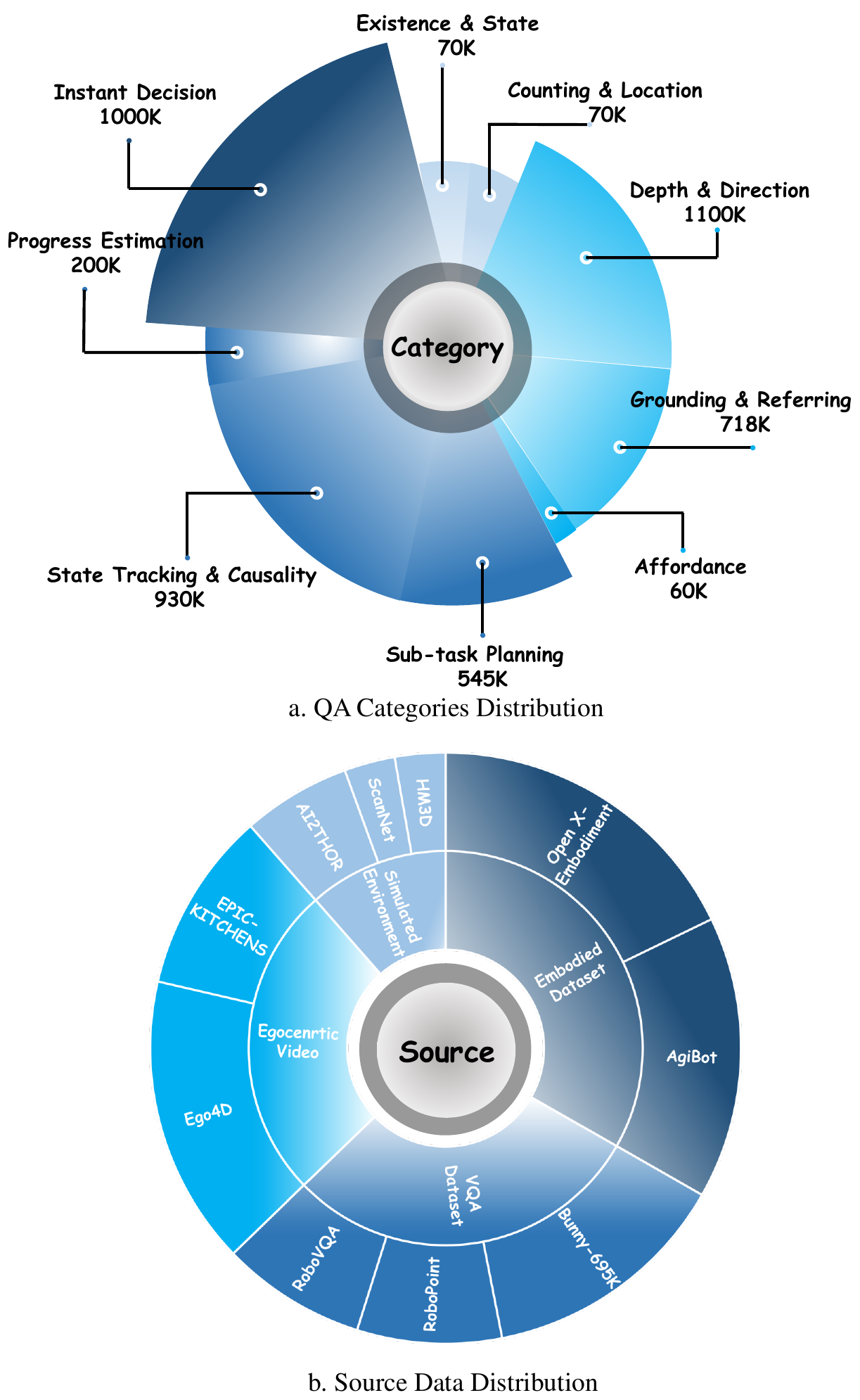}
    
    \caption{QA categories distribution and source data distribution in the EQA-Decision dataset.}
    \label{fig:data_distribution}
\end{figure}

Figure~\ref{fig:data_distribution} presents the overall statistics of EQA-Decision.  
The dataset contains over four million multimodal question-answer pairs across nine reasoning sub-tasks.  
Among them, depth and direction (1.1M), state tracking and causality (0.93M), and instant decision (1.0M) constitute the major portions, highlighting the focus on spatial, temporal, and decision-oriented reasoning.

EQA-Decision integrates four complementary data sources: simulated environments, image-QA datasets, egocentric videos, and embodied robot trajectories.  
The data are collected from diverse embodied contexts that combine simulated, visual, and real-world sources.  
This integration ensures both large-scale diversity and temporal realism, providing balanced coverage across perception, reasoning, and decision-making dimensions.  
Detailed statistical analysis is provided in the supplementary material.

\section{RoboDecision}
\label{sec:baseline}

\subsection{Model Base}
Our baseline is Qwen3-VL-8B, which follows a vision encoder and language decoder architecture. 
It employs a DeepStack fusion mechanism, where multilevel ViT features are repeatedly injected into several transformer layers of the language model to achieve fine-grained multimodal alignment. 
The vision encoder processes native-resolution images or video frames using multi-resolution rotary positional encoding (MRoPE) to capture spatial and temporal cues. 
The hierarchically merged visual tokens are passed to a decoder-only transformer that supports long-context reasoning, providing a strong foundation for spatio-temporal reasoning in embodied intelligence.
\subsection{Training Strategies}
As illustrated in Figure~\ref{fig:main}, 
we adopt a three-stage training pipeline that progressively enhances multimodal reasoning and decision-making ability.

\vspace{-4mm}
\paragraph{Stage 1: Supervised Fine-Tuning (SFT)}  
The model is initialized from Qwen3-VL-8B-Instruct and trained on the EQA-Decision dataset using LoRA~\cite{hu2022lora}. 
During this stage, the vision encoder remains frozen for stability, while the language and fusion layers are optimized to improve cross-modal alignment. 
Training data are uniformly sampled from all four reasoning modules to ensure balanced coverage across perceptual, spatial, task, and decision-oriented contexts. 
This stage focuses on injecting embodied-domain knowledge and enhancing the model’s basic spatial, temporal, and decision reasoning abilities, establishing a foundation for subsequent stages. 

\vspace{-4mm}
\paragraph{Stage 2: Chain-of-Thought Supervised Fine-Tuning (CoT-SFT)} Based on the SFT foundation, this stage enhances the model’s reasoning depth and interpretability through structured chain-of-thought supervision.  
We uniformly sample about 10\% of the training data across all modules and employ Gemini-2.5-pro~\cite{comanici2025gemini} to generate CoT annotations that contain both rationale and answer fields.  

We then finetune the model exclusively on this CoT-augmented subset using the same LoRA~\cite{hu2022lora} configuration and optimizer as in Stage~1.
Through this process, the model learns to construct coherent multi-step reasoning chains across spatial and temporal contexts, forming a more explicit understanding of causal dependencies within embodied tasks.  In addition, reasoning-aware supervision stabilizes reward signals and provides a warm initialization for subsequent GRPO~\cite{shao2024deepseekmath} optimization.
\begin{table*}[t]
\centering
\small
\setlength{\tabcolsep}{4pt}
\renewcommand{\arraystretch}{1.15}
\begin{adjustbox}{max width=\textwidth}
\begin{tabular}{l|cccccc|c}
\hline
\multirow{2}{*}{\textbf{Models}} & \multicolumn{6}{c|}{\textbf{EQA-Decision Benchmark}} & \multirow{2}{*}{\textbf{Overall }} \\
& Static Scene & Spatial–Depth & Grounding–Referring & Temporal Reasoning & Planning Reasoning & Instant Decision & \\
\hline
\multicolumn{8}{l}{\textbf{General Baselines}} \\
\hline
Gemini-2.5-Pro & 66.70 & 56.54 & 47.56 & 17.42  & 56.35 & 47.56 & 48.68 \\
GPT-5 & 70.74  & 47.75  & 45.25 & 25.72 & 54.52 & 62.25 & 51.03  \\
Qwen3-VL-4B-Thinking & 76.80 & 51.75 & 15.36 & 15.02 & 48.70 & 55.27 & 43.81\\
Qwen3-VL-8B-Thinking & 76.65 & 55.42 & 23.14 & 16.72 & 51.75  & 55.75 & 46.57 \\
Qwen3-VL-4B-Instruct & 74.23 & 51.97 & 20.19 & 15.83 & 53.35 & 52.36 & 44.65 \\
Qwen3-VL-8B-Instruct & 70.47 & 54.84 & 35.51 & 23.98 & 54.02 & 54.27 &  48.84 \\
\hline
\multicolumn{8}{l}{\textbf{Embodied Baselines}} \\
\hline
RoboBrain-7B-2.0 & 60.80 & 25.62  & 61.93 & 19.25  & 33.70  & 41.90 & 37.32 \\
\rowcolor{cyan!10}
\textbf{RoboDecision-8B} & \textbf{81.55} & \textbf{70.82} & \textbf{68.12}&\textbf{52.95} & \textbf{65.02} & \textbf{69.93} & \textbf{68.06} \\
\hline
\end{tabular}
\end{adjustbox}
\caption{Comprehensive results on our benchmark with six task modules. The highlighted row denotes RoboDecision model.}
\label{tab:overall_bench}
\end{table*}

\vspace{-4mm}
\paragraph{Stage 3: Reinforcement Fine-Tuning (GRPO)}
Following the CoT-SFT stage, we introduce a reinforcement phase based on Group Relative Policy Optimization (GRPO)~\cite{shao2024deepseekmath} to further strengthen the model's ability to connect perceptual understanding with decision-making.
Although SFT establishes solid language alignment, the model often relies on textual priors and may produce similar outputs even when visual contexts differ.
GRPO mitigates this issue by guiding learning through a hybrid reward that explicitly encourages image-grounded reasoning, addressing a critical gap along the perception-to-decision path.

The hybrid reward integrates reasoning quality, answer correctness, and visual consistency:
\begin{eqnarray}
R_{\text{total}} = \alpha R_{\text{reason}} + \beta R_{\text{answer}} + \gamma R_{\text{visual}},
\end{eqnarray}
where $\alpha, \beta, \gamma$ are adaptive weights that vary with task type.
The reasoning reward evaluates the coherence and causal soundness of the generated rationale.
We embed both the predicted reasoning trace and the reference chain-of-thought using an E5-large encoder~\cite{wang2024multilingual} and compute their cosine similarity, encouraging structured reasoning that reflects causal relations, spatial justification, and temporal dependencies derived from the scene.

The answer reward measures factual correctness. For free-form responses, we compute semantic similarity using E5-large~\cite{wang2024multilingual} embeddings.
For structured decision outputs including spatial coordinates, bounding boxes, or depth values, we apply task-specific rule-based scoring functions. The visual consistency reward aligns the reasoning process with the input image. We compute the similarity between an OpenCLIP~\cite{cherti2023reproducible} embedding of the visual observation and an embedding of the generated rationale.
High scores are obtained only when the reasoning accurately reflects what is visually present, ensuring that the model's thought process remains anchored to perceptual evidence rather than drifting toward textual bias.

By jointly optimizing these signals, the model shifts from a text-driven responder to a perception-guided decision maker. 
Rather than repeating QA patterns, it learns to reason directly from the image and generate decisions that adapt to spatial layout, scene dynamics, and task demands.
Additional training details and implementation settings are provided in the supplementary material.

\section{Experiments}
\subsection{EQA-Decision Benchmark}

To systematically evaluate the key abilities required for embodied intelligence,
we construct the \textbf{EQA-Decision Benchmark}, a unified evaluation suite designed
to assess perception, reasoning, and decision-making in dynamic embodied settings.
Each dimension adopts task formulations aligned with the annotation procedures in our dataset, 
while all benchmark samples remain strictly disjoint from the training set, ensuring 
clean separation between training and evaluation.

The benchmark covers six task categories aligned with the evaluation
dimensions: static scene understanding,
spatial-depth reasoning, grounding and referring, temporal reasoning,
planning reasoning, and instant decision-making.
It contains a total of 2,118 carefully curated evaluation samples,
including 264 items for static scene understanding, 314 for spatial-depth reasoning,
200 for grounding-referring, 338 for temporal reasoning, 480 for planning reasoning,
and 522 for instant decision evaluation.
All benchmark items follow a unified multimodal QA format with balanced category distribution, providing a consistent and fair basis for evaluating perception, reasoning, and decision-making capabilities.

\begin{table}[t]
\centering
\small
\setlength{\tabcolsep}{4pt}
\renewcommand{\arraystretch}{1.2}

\begin{adjustbox}{max width=\linewidth}
\begin{tabular}{l|cccc|c|c}
\hline
\multirow{2}{*}{\textbf{Models / Metrics} }
    & \multicolumn{4}{c|}{\textbf{RoboVQA}} 
    & \textbf{ERQA} 
    & \textbf{Where2Place} \\
 & BLEU-1 & BLEU-2 & BLEU-3 & BLEU-4
 & All  
 & All \\
\hline
\multicolumn{7}{l}{\textbf{General Baselines}} \\
\hline
Gemini-2.5-Pro  & 34.76 & 30.82 & 29.97 &23.63  & 48.70  &  18.11 \\
GPT-5 &36.48 & 27.75 &25.73  & 24.92 & 49.95 & 25.58 \\
Qwen3-VL-8B-Instruct & 82.34 & 69.82 & 61.25 & 18.64 & 42.50  & 32.45 \\
\hline
\multicolumn{7}{l}{\textbf{Embodied Baselines}} \\
\hline

RoboBrain-7B-2.0 &49.56 &33.83 & 26.59 & 12.20 & 39.44  & 63.59 \\

\rowcolor{cyan!15}
\textbf{RoboDecision-8B} 
&\textbf{86.97}  
     &  \textbf{80.82}
    & \textbf{72.23} 
    & \textbf{43.55} 
    & \textbf{54.50} 
    & \textbf{67.08} \\
\hline
\end{tabular}
\end{adjustbox}

\caption{Performance comparison across benchmarks. The highlighted row denotes RoboDecision model.}
\label{tab:otherbenchmark}
\end{table}

We adopt task-specific metrics tailored to the heterogeneous nature of embodied reasoning. 
For grounding and referring tasks, evaluation is conducted in pixel space.
Model outputs are interpreted as either 2D points or rectangular regions, with
normalized coordinates converted to the image resolution. Rectangular predictions
are expanded into dense pixel sets so that all outputs are handled uniformly.
These predicted locations are compared against the ground-truth binary mask, and
accuracy is computed as the average mask value over all valid predicted points,
reflecting their spatial overlap with the annotated target region.
For all other QA types, we employ LLM-Match, where GPT-5 assigns each response a correctness score on a 1–5 scale based on its agreement with the ground-truth answer. The score is then linearly mapped to the [0,100] range and averaged across all samples, providing a unified evaluation metric applicable to all question types.
\begin{table}[t]
\centering
\small
\setlength{\tabcolsep}{3pt}
\renewcommand{\arraystretch}{1.12}

\begin{adjustbox}{max width=\linewidth}
\begin{tabular}{l|ccccccc}
\hline
\textbf{Module} 
    & \textbf{Full}
    & \textbf{w/o GRPO}
    & \textbf{w/o CoT}
    & \textbf{w/o Scene}
    & \textbf{w/o Spatial}
    & \textbf{w/o Task}
    & \textbf{w/o Decision} \\
\hline
Static Scene        & \textbf{81.55} & 80.23 & 79.37 & 75.36 & 80.18 & 81.33 & 81.71 \\
Spatial--Depth      & \textbf{70.82} & 60.91 & 58.34 & 69.23 & 58.17 & 59.52 & 71.74 \\
Grounding--Referring& \textbf{68.12} & 55.63 & 44.27 & 68.51 & 41.72 & 66.07 & 67.38 \\
Temporal Reasoning  & \textbf{52.95} & 45.18 & 31.46 & 51.37 & 50.92 & 42.74 & 51.22 \\
Planning Reasoning  & \textbf{65.02} & 57.74 & 56.38 & 65.24 & 63.84 & 56.53 & 60.11 \\
Instant Decision    & \textbf{69.93} & 59.44 & 57.33 & 67.41 & 60.14 & 58.27 & 55.84 \\
\hline
\textbf{Overall}    & \textbf{68.06} & 59.85 & 54.52 & 66.18 & 59.16 & 60.74 & 64.61 \\
\hline
\end{tabular}
\end{adjustbox}

\caption{
Ablation study across training stages (GRPO / CoT-SFT) and data modules (Scene, Spatial, Task, Decision).
}
\label{tab:ablation_single_column}
\end{table}

\subsection{Main Results}
We evaluate a wide variety of multimodal systems on the EQA-Decision Benchmark with results summarized in Table~\ref{tab:overall_bench}. 
The evaluated models include Gemini-2.5-Pro~\cite{comanici2025gemini}, GPT-5, Qwen3-VL-4B-Thinking, 
Qwen3-VL-8B-Thinking, Qwen3-VL-4B-Instruct, Qwen3-VL-8B-Instruct, 
RoboBrain-7B-2.0~\cite{team2025robobrain}, and our proposed RoboDecision-8B. 
These models span frontier proprietary VLMs, leading open-source MLLMs, 
and embodied-robotics systems specifically optimized for spatial reasoning and manipulation tasks.

RoboDecision-8B achieves the highest overall score of \textbf{68.06}, establishing state-of-the-art performance across all six task modules on the EQA-Decision Benchmark.
These results demonstrate the model's ability to integrate spatial cues, temporal progression, 
and action-level reasoning---capabilities essential for decision-oriented embodied intelligence.
RoboDecision-8B consistently outperforms both open-source and embodied-robotics
baselines across our benchmark. Compared with Qwen3-VL-8B-Instruct, the overall
score improves from 48.84 to 68.06, with the largest gains in grounding--referring,
temporal stage identification, and planning reasoning. These improvements result 
from our structured CoT supervision and visual--semantic alignment rewards, which 
encourage the model to ground its reasoning in the visual scene rather than rely on 
memorized patterns. RoboDecision-8B also surpasses RoboBrain-7B-2.0~\cite{team2025robobrain}, whose overall 
score is 37.32, particularly in spatial--depth understanding and multi-step planning. 
This demonstrates that our multi-stage training pipeline strengthens visual grounding and perception-to-decision reasoning, achieving robust and generalizable performance across embodied QA tasks. 

To further validate generality, we also evaluate RoboDecision-8B on three representative embodied benchmarks: RoboVQA~\cite{sermanet2024robovqa} for long-horizon robot VQA, ERQA~\cite{team2025gemini} for fine-grained embodied reasoning, and Where2Place~\cite{yuan2024robopoint} for free-space placement understanding. As shown in Table~\ref{tab:otherbenchmark}, RoboDecision-8B consistently outperforms both general-purpose VLMs and embodied baselines across all tasks. These results show that the model’s capabilities transfer effectively beyond EQA-Decision to other embodied benchmarks.
\subsection{Ablation Studies}

To quantify the contribution of each component in our training pipeline and dataset, we conduct a set of controlled ablations, with results summarized in Table~\ref{tab:ablation_single_column}.

\textbf{Investigation on training stages}
Ablation results confirm the complementary and necessary roles of CoT-SFT and GRPO. CoT-SFT is crucial for multi-step reasoning, and its removal causes a severe performance drop, particularly in grounding and temporal reasoning. In contrast, GRPO focuses on optimizing decisions, and its removal primarily weakens spatial understanding, grounding quality, and instant decision-making.

\textbf{Investigation on data modules}
We analyze the contribution of each data module by separately removing the four components. Removing Spatial data weakens spatial skills, removing Task data reduces temporal reasoning, and excluding Decision data most strongly harms Instant Decision. Introducing Scene data mainly benefits scene-centric perception and yields modest gains in Instant Decision and Temporal Reasoning, while exerting  mild influence on other modules.

Instant Decision also shows clear cross-module sensitivity, dropping when Spatial or Task data is ablated, indicating that effective decisions require combining spatial layout, temporal cues, and task progression. Other modules show more localized degradation. Overall, the four modules provide complementary supervision, and robust perception-to-decision behavior emerges only when they are jointly used.
\begin{figure}[t]
    \centering
    \includegraphics[width=1\linewidth]{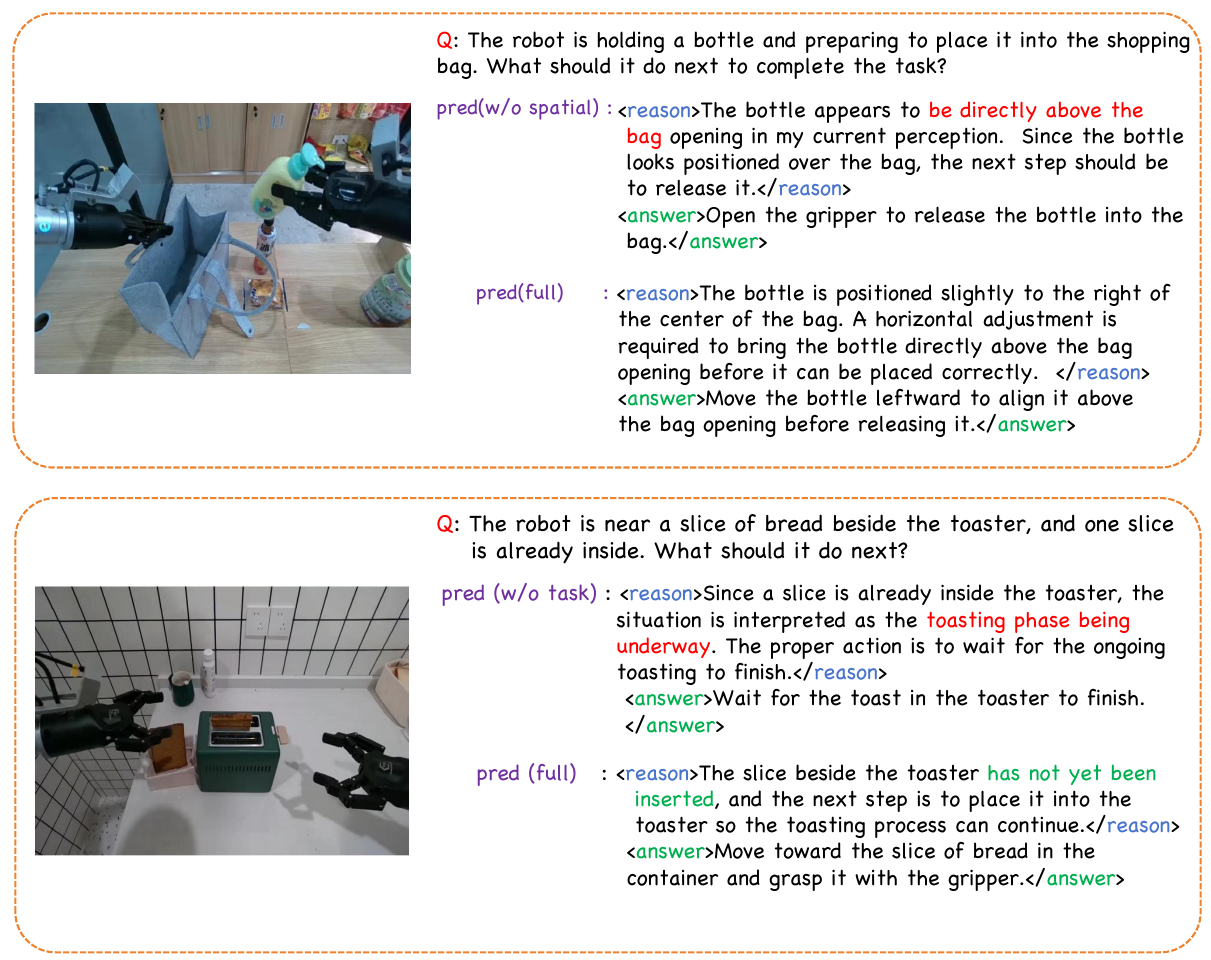}
    \caption{Qualitative comparison of instant decisions.}
    \label{fig:toast_spatial}
\end{figure}


\subsection{Qualitative Analysis}

Our qualitative results (Fig.~\ref{fig:toast_spatial}) show that the full model produces
consistent decisions by jointly using spatial cues, temporal progress, and task intent, 
while ablated variants tend to rely on partial evidence and make incorrect choices. 
When spatial data is removed, the model misjudges object positions and issues premature 
actions; when task-dynamics data is removed, it misinterprets the stage of the ongoing 
task and selects inappropriate next steps. Overall, these examples highlight the same 
dependency patterns observed in our quantitative ablations: reliable perception-to-decision 
behavior emerges only when spatial understanding, temporal reasoning, and task-aware 
interpretation are learned together.

\section{Conclusion}

We presented EQA-Decision, a large-scale dataset that extends embodied
question answering from static perception to decision-centric reasoning through
structured supervision spanning spatial understanding, task dynamics, temporal
progression, and instant decisions. Building on this dataset, we developed the
RoboDecision framework, which integrates SFT, CoT-SFT, and GRPO with a
hybrid reward to align reasoning, answers, and visual grounding. Experimental
results show that RoboDecision-8B achieves state-of-the-art performance across
six embodied reasoning types, while ablation studies confirm the complementary
roles of both the data modules and training stages. Together, EQA-Decision and
RoboDecision provide a solid foundation for advancing decision-oriented embodied
intelligence.

\textbf{Acknowledgment}: This work was supported in part by research resources provided through the collaboration between Peking University and XYZ Embodied AI.

{
    \small
    \bibliographystyle{ieeenat_fullname}
    \bibliography{main}
}
\clearpage

\end{document}